\documentclass{article}

\usepackage{microtype}
\usepackage{graphicx}
\usepackage{subfigure}
\usepackage{booktabs} 

\usepackage{hyperref}

\usepackage{multirow}

\newcommand{\tabincell}[2]{\begin{tabular}{@{}#1@{}}#2\end{tabular}} 

\usepackage[accepted]{icml2024}

\usepackage{amsmath}
\usepackage{amssymb}
\usepackage{mathtools}
\usepackage{amsthm}

\usepackage[capitalize,noabbrev]{cleveref}

\theoremstyle{plain}
\newtheorem{theorem}{Theorem}[section]

\theoremstyle{definition}
\newtheorem{definition}[theorem]{Definition}

\theoremstyle{remark}

\usepackage[textsize=tiny]{todonotes}

\icmltitlerunning{Compositional Few-Shot Class-Incremental Learning}

\begin{document}

\twocolumn[
\icmltitle{Compositional Few-Shot Class-Incremental Learning}



\icmlsetsymbol{equal}{*}

\begin{icmlauthorlist}
\icmlauthor{Yixiong Zou}{hust}
\icmlauthor{Shanghang Zhang}{pku}
\icmlauthor{Haichen Zhou}{hust}
\icmlauthor{Yuhua Li}{hust}
\icmlauthor{Ruixuan Li}{hust}

\end{icmlauthorlist}

\icmlaffiliation{hust}{School of Computer Science and Technology, Huazhong University of Science and Technology, Wuhan, China}
\icmlaffiliation{pku}{School of Computer Science, Peking University, Beijing, China}

\icmlcorrespondingauthor{Ruixuan Li}{rxli@hust.edu.cn}

\icmlkeywords{Few-shot class-incremental learning, Compositional learning}

\vskip 0.3in
]



\printAffiliationsAndNotice{}

\begin{abstract}
Few-shot class-incremental learning (FSCIL) is proposed to continually learn from novel classes with only a few samples after the (pre-)training on base classes with sufficient data. However, this remains a challenge. In contrast, humans can easily recognize novel classes with a few samples. Cognitive science demonstrates that an important component of such human capability is compositional learning. This involves identifying visual primitives from learned knowledge and then composing new concepts using these transferred primitives, making incremental learning both effective and interpretable. 
To imitate human compositional learning, we propose a cognitive-inspired method for the FSCIL task. We define and build a compositional model based on set similarities, and then equip it with a primitive composition module and a primitive reuse module. In the primitive composition module, we propose to utilize the Centered Kernel Alignment (CKA) similarity to approximate the similarity between primitive sets, allowing the training and evaluation based on primitive compositions. In the primitive reuse module, we enhance primitive reusability by classifying inputs based on primitives replaced with the closest primitives from other classes. Experiments on three datasets validate our method, showing it outperforms current state-of-the-art methods with improved interpretability. 
Our code is available at https://github.com/Zoilsen/Comp-FSCIL.
\end{abstract}

\section{Introduction}

With advancements in hardware, deep neural networks have demonstrated considerable success across various areas using pre-defined large-scale datasets~\cite {simonyan2014very,he2016deep}. However, real-world scenarios present novel knowledge continuously, often with limited data~\cite{hou2019learning,rebuffi2017icarl}, such as rare diseases. Addressing this challenge requires models to learn novel knowledge from just a few samples without forgetting previously learned ones~\cite{castro2018end,tao2020few}. This necessity gives rise to the Few-Shot Class-Incremental Learning (FSCIL) task~\cite{zhang2021few,zhou2022forward}. In this task, models are initially (pre-)trained on base classes during a base session with sufficient training data. Then, models learn from novel classes in incremental sessions with only a few samples, and finally classify test samples across all encountered classes. While various approaches, including metric-based ones~\cite{zou2022margin,zhang2021few}, adaptation-based ones~\cite{zhou2022forward,Yang2023Neural}, etc. have been explored for this task, FSCIL remains a challenge due to the scarcity of training data and the risk of catastrophic forgetting.

\begin{figure}[t]
	\centering
	\includegraphics[width=1.0\linewidth]{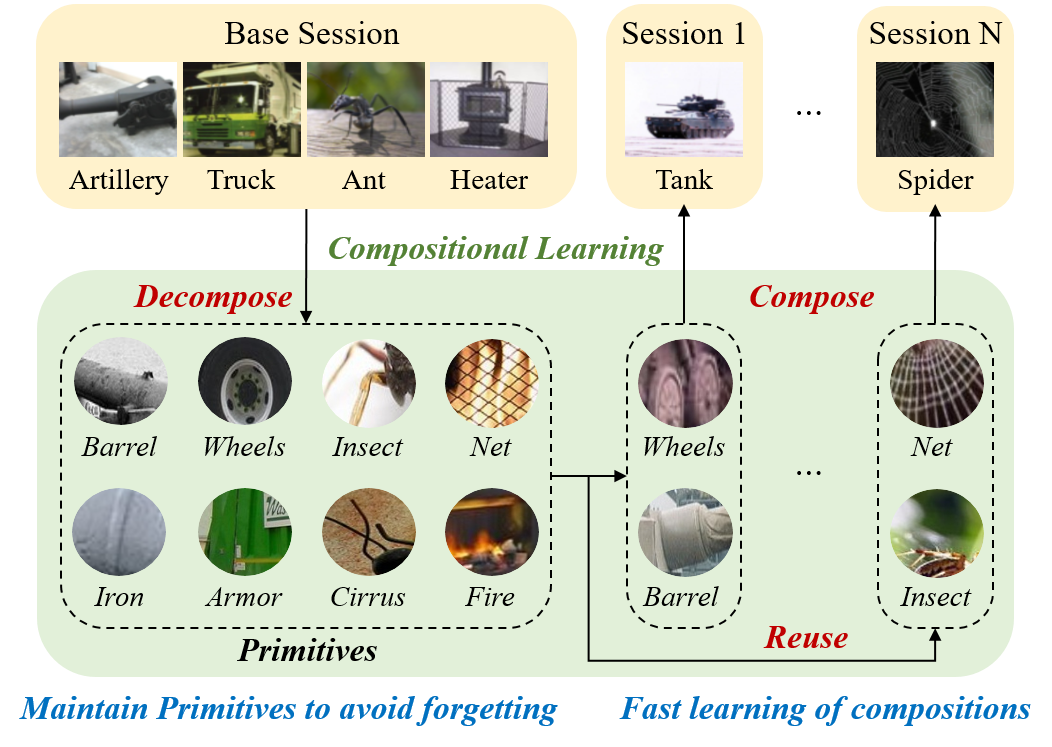}\vspace{-0.4cm}
	\caption{
		As studied by cognitive science~\cite{biederman1987recognition}, humans can compositionally learn knowledge by dividing learned ones into primitives, and then compose them to learn novel knowledge, which leads to the good human capability of incremental learning with only scarce data. To imitate the human ability of compositional learning, we propose a compositional learning method for the few-shot class-incremental learning (FSCIL) task. We briefly plot the primitives automatically found by our methods with the possible meanings, where we can see good reusability and interpretability of primitives. 
		Detailed plots are in Fig.\ref{fig:composition}.
	}
	\label{fig:motivation}
\end{figure}

In contrast to machines, humans can easily learn from limited data without forgetting learned knowledge~\cite{DBLP:journals/corr/abs-1906-01905}. Cognitive science demonstrates that an important component of such human capability is compositional learning~\cite{biederman1987recognition,zou2020compositional}, which enables humans to divide knowledge, such as semantic objects, into visual primitives~\cite{hoffman1984parts} (like object parts), and then compose novel or learned knowledge by transferred primitives~\cite{fodor1975language}, as shown in Fig.\ref{fig:motivation}. 
Since primitives are reusable among base and novel classes, 
it enables us to not only avoid forgetting by maintaining learned primitives, but also efficiently learn from few-shot novel classes by learning the composition of primitives.
Moreover, primitives can be viewed as the foundational elements guiding human decision-making, providing insights into why a particular sample is classified into a specific class. This enhances the interpretability of black-box deep learning models (e.g., Fig.\ref{fig:motivation} depicts a spider as a composition of an ant-like insect and nets). Consequently, this paper aims to imitate the human ability for compositional learning to tackle the challenging problem of FSCIL. 

Specifically, we first define primitive composition based on set similarities and then build our model by modifying the FSCIL base method. Since primitives always refer to object parts~\cite{zou2020compositional}, we define image patches as candidate primitives, which may contain sample-specific candidate primitives such as background, and common primitives shared across samples. For each class, we employ a set of prototypes to form its primitive set, which encodes common primitives shared within this class. We then propose to utilize the Centered Kernel Alignment (CKA) similarity~\cite{kornblith2019similarity} to approximate the similarity between primitive sets, which enables the training and evaluation based on primitive compositions. To enhance the reusability of primitives across classes, we further design a primitive reuse module, which classifies input samples based on primitives replaced with the closest primitives from other classes. Our model is firstly trained in the base session, and then transferred to incremental sessions with a fixed backbone network. The reusability of primitives is achieved both implicitly through the reuse of the backbone network and explicitly through the primitive reuse module.

In summary, our contributions can be listed as:

$\bullet$ We propose a cognitive-inspired compositional learning method for the FSCIL task, which first defines and builds a compositional model based on set similarities, and then equips it with a primitive composition module and a primitive reuse module.

$\bullet$ In the primitive composition module, we propose to utilize the CKA similarity to approximate the similarity between primitive sets, allowing the training and evaluation based on primitive compositions.

$\bullet$ In the primitive reuse module, we enhance primitive reusability by classifying inputs based on primitives that are replaced with the nearest primitives from other classes.

$\bullet$ Extensive experiments on three public benchmarks validate the rationale of our compositional learning method, and demonstrate it outperforms current state-of-the-art works while providing enhanced interpretability.

\section{Related Work}

\textbf{Few-shot class-incremental learning} can be roughly categorized into adaptation-based~\cite{hou2019learning,rebuffi2017icarl,castro2018end,tao2020few} and metric-based methods~\cite{zhang2021few,zou2022margin}. The first group adapts the model during novel-class training, with the backbone network often frozen to prevent catastrophic forgetting~\cite{zhou2022forward}. In the second group, each class is represented by prototypes averaged from samples~\cite{zou2022margin}, with network parameters similarly frozen to mitigate the risk of catastrophic forgetting. However, there is a scarcity of works exploring the compositional structure of FSCIL models, and as far as we know, our study is the first to delve into this aspect.

\textbf{Compositional learning} seeks to learn knowledge through its primitives or components, which is a concept extensively explored in cognitive science~\cite{biederman1987recognition, hoffman1984parts, fodor1975language}. This approach has found applications in various domains, such as CPDE~\cite{zou2020compositional} decomposing classes into channels for few-shot learning, \cite{purushwalkam2019task} breaking down visual features into attributes for zero-shot learning, \cite{kato2018compositional} decomposing human-object interactions into actions and objects, \cite{cao2021concept} learning a dictionary for visual concepts, and \cite{tang2020revisiting} aligning object parts with pose normalization. However, there has been limited exploration in the FSCIL task. In contrast, our decomposition operates within the spatial dimension and does not need additional annotations for primitives. Due to space limitations, we present further details on related works in the appendix.

\begin{figure*}[t]
	\centering
	\includegraphics[width=1.0\linewidth]{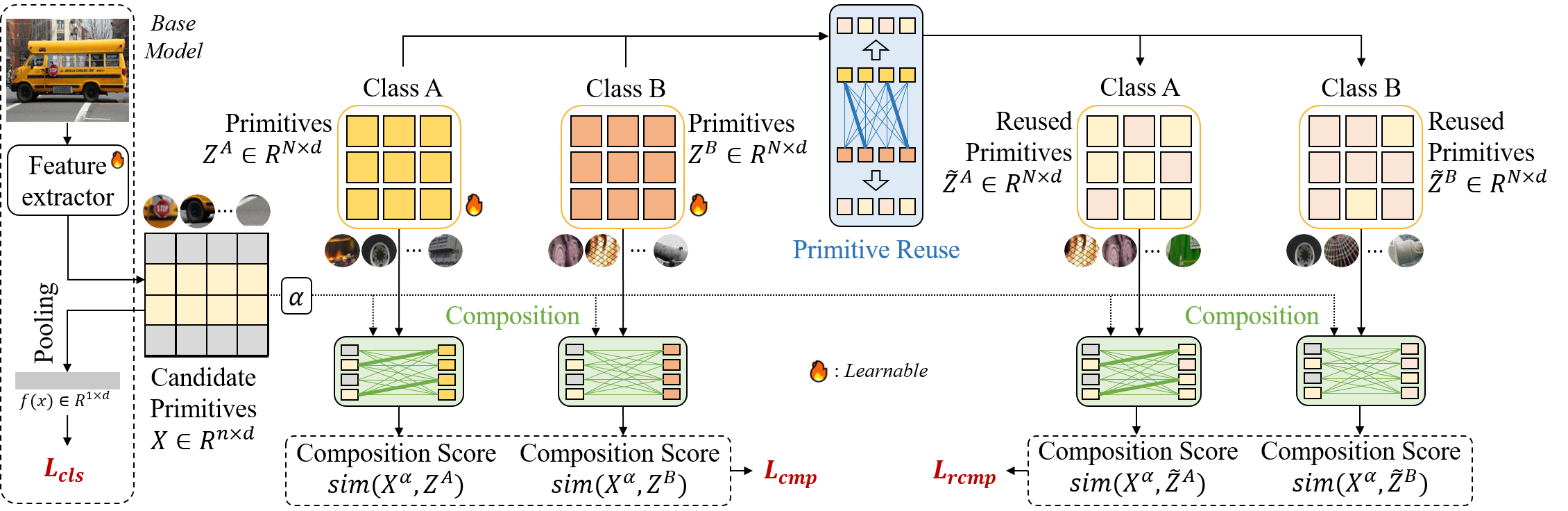}\vspace{-0.3cm}
	\caption{
		We take image patches as candidate primitives, and utilize a set of prototypes to construct the primitive set for each class. 
		Given an input sample, our method tries to compose it with primitive sets (e.g., $Z^A$ and $Z^B$) from different classes (e.g., class A and B), which is measured as the composition score by the CKA similarity. 
		These composition scores are then utilized to be the classification score for the model training and evaluation. 
		To improve the reusability of primitives across classes, each primitive is replaced with the closest primitive in other classes. The replaced primitive sets (e.g., $\tilde{Z}^A$ and $\tilde{Z}^B$) will then be applied in the classification with the composition score. 
		Finally, our model is trained with the combination of $L_{cls}$, $L_{cmp}$, and $L_{rcmp}$ during both the base and incremental sessions.
	}
	\label{fig:framework}
\end{figure*}

\section{Method}

We first define primitive composition by set similarities and then design each component to implement compositional few-shot class-incremental learning (FSCIL) (Fig.\ref{fig:framework}).

\subsection{Preliminaries}

FSCIL~\cite{zhang2021few,zhou2022forward} aims to continually learn from novel classes with only a few samples in incremental sessions, after (pre-)training on base classes with abundant training data in the base session. Initially, the model is trained on the base session dataset $D^0 = \{(x_i, y_i)\}_{i=1}^{n_0}$ with the label space $Y_0$, by minimizing the loss $\sum_{(x_i, y_i) \in D^0} L(\phi(W, x_i), y_i)$, where $L(\cdot, \cdot)$ denotes the cross-entropy loss and $\phi(\cdot, \cdot)$ gets the prediction of $x_i$. Typically, $\phi(\cdot, \cdot)$ consists of a feature extractor $f(\cdot)$ and a classifier (e.g., fully connected (FC) layer) with the parameter $W$, where $\phi(W, x) \in R^{1 \times |Y_0|}$, $W \in R^{|Y_0| \times d}$ and $f(x) \in R^{1 \times d}$. During the $k$th incremental session, the model learns from the dataset $D^k = \{(x_i, y_i)\}_{i=1}^{n_k}$ with the label space $Y_k$. The classifier's weight $W$ will be extended by incorporating the classifier obtained from $D^k$ as $W = \{w^0_1, w^0_2, ..., w^0_{|Y_0|}\} \cup ... \cup \{w^k_1, ..., w^k_{|Y_k|}\} \in R^{\sum_{i=0}^{k} |Y_i| \times d}$ where $w^k_j \in R^{1 \times d}$ denotes the classifier weight for the $j$th class in the $k$th session. A prevailing baseline method~\cite{zhang2021few} freezes $f(\cdot)$ during incremental sessions and only trains the classifier. Finally, the model will be applied to classify test samples from all encountered $\sum_{i=0}^{k} |Y_i|$ classes.


\subsection{Defining the Compositional Recognition}


Humans' compositional learning first divides knowledge into primitives~\cite{hoffman1984parts,zou2020compositional}, and then composes novel knowledge using these primitives~\cite{fodor1975language}. 
This learning mechanism avoids forgetting by maintaining learned primitives, and facilitates few-shot learning by efficiently learning the composition of reused primitives.
To imitate this human ability, we begin with the following definition.
\begin{definition}
	Given a class $y$, compositional learning divides it into a set of primitives $\{P^y_i\}_i^N$, representing shared components in this class. Each sample $x$ is divided into a set of components $\{C_i(x)\}_i^n$, where components shared with other samples in this class construct $\{P^y_i\}_i^N$, while other components are specific to this sample.
	\label{def: primitive}
\end{definition}

Therefore, we refer to ${C_i(x)}_i^n$ as the candidate primitive set. We use the notations $Z^y$ and $X$ as abbreviations for $\{P^y_i\}_i^N$ and ${C_i(x)}_i^n$, respectively.
Note that primitives should be transferable or even reusable across classes.
Considering that each sample and class are represented by sets, if all elements in set $X$ are present in set $Z^y$, we can assert that $X$ is composed of elements from $Z^y$. Therefore, we define composition using the Jaccard Similarity as:
\begin{definition}
	Set $X = \{C_i(x)\}_i^n$ is composed of elements from set $Z^y = \{P^y_i\}_i^N$ if $\frac{|Z^y \cap X|}{|Z^y \cup X|} = 1.0$.
\end{definition}

However, this criterion is strict and hard to apply due to two reasons: (1) the term $|Z^y \cap X|$ is not continuous, and (2) achieving  $\frac{|Z^y \cap X|}{|Z^y \cup X|} = 1.0$ is difficult because $X$ may include elements specific to the sample (e.g., background). 

Therefore, we relax this criterion to $\frac{|Z^y \cap X|}{|Z^y \cup X|} > t$. Considering the sparsity of primitives~\cite{zou2020compositional}, we assume $|Z^y| << |X|$ and simplify the Jaccard Similarity as

\vspace{-0.5cm}
{\small
	\begin{align}
		\frac{|Z^y \cap X|}{|Z^y \cup X|} = \frac{|Z^y \cap X|}{|X| + |Z^y| - |Z^y \cap X|} \approx \frac{|Z^y \cap X|}{|X|},
		\label{eq:trans_jaccard}
\end{align}}since $|X| >> |Z^y| > |Z^y| - |Z^y \cap X|$. Moreover, we relax the concrete union of sets to the similarity between sets as

\vspace{-0.5cm}
{\small
	\begin{align}
		|Z^y \cap X| \approx sim(X, Z^y) = \sum\nolimits_i^{|X|} \sum\nolimits_j^{|Z^y|} s(X_i, Z^y_j),
		\label{eq:set_similarity}
\end{align}}where $s$ denotes the similarity between primitives. Ideally, $s(\cdot,\cdot)$ outputs 1 if $X_i$ could totally match $Z^y_i$, and 0 if $X_j$ and $Z^y_i$ are not matched (e.g., cosine similarity).
Since $sim(X, Z^y)$ is continuous, we can utilize it to represent the probability of classifying the sample $x$ into the class $y$ as

\vspace{-0.5cm}
{\small
	\begin{align}
		P(y|x) = \frac{e^{\tau \cdot sim(X, Z^y) / |X|}}{\sum_k e^{\tau \cdot sim(X, Z^k) / |X|}}.
		\label{eq:compose_to_recognize}
\end{align}}where $\tau$ is a temperature parameter. Naturally, $P(y|x)$ can be used in training and evaluation.
Therefore, there remain three issues to implement the compositional learning: (1) designing the primitives; (2) designing the set similarity function $sim(\cdot, \cdot)$ and (3) designing the reuse of primitives.

\subsection{(Candidate) Primitive Design}

To achieve this goal, we look back into the FSCIL base method (Fig.\ref{fig:framework}).
Given an input image $x$, this model extracts its feature as $f(x) \in R^{1 \times d}$, and then forwards it to the classifier with the parameter $W \in R^{|Y_0| \times d}$, 
where $W_y \in R^{1 \times d}$ is viewed as the prototype of the $y$th class. Typically, these features have been processed by the Global-Average-Pooling layer, such as ResNet~\cite{he2016deep} and Swin Transformer~\cite{liu2021Swin}. Therefore, we have $f(x) = \frac{1}{S} \sum_{i}^{S} F(x)_i$, where $F(x) \in R^{S \times d}$ denotes the feature map and $S$ is the spatial dimension of the map. Given that the object in the input image is partitioned into image patches, the patch features $F(x)$ can be seen as compositionally representing the input $x$. Therefore, it can be regarded as the candidate primitive set containing object parts, i.e.,

\vspace{-0.5cm}
{\small
	\begin{align}
		X = \{C_i(x)\}_i^n = \{F(x)_i\}_i^S.
		\label{eq:patch_to_primitive}
\end{align}}Furthermore, this design satisfies the transferability requirements of primitives, as patch features are more readily transferable compared with image features.
With this design, the term $|X|$ in Eq.\ref{eq:compose_to_recognize} is a constant number $S$. Therefore, the designing of the compositional model is simplified into finding a suitable similarity function between $X$ and $Z^y$.

On the other hand, since $W_y$ inherently learns to represent the centroid of class $y$ during the training of the base model, it captures the shared patterns of the class $y$ while disregarding the sample-specific patterns like the background.
Similarly, in line with Eq.\ref{eq:patch_to_primitive}, we replace $W_y$ of class $y$ with a collection of prototypes to be the primitive set $\{P^y_i\}_i^N$ where $P^y_i \in R^{1 \times d}$. 
With this choice of primitive set, similarly, $\{P^y_i\}_i^N$ would also learn the common image patches (i.e., candidate primitives) that are shared among different samples from the class $y$, which serves as the centroid of candidate primitives and ignores sample-specific candidate primitives such as background. Therefore, this choice of primitive set satisfies the definition~\ref{def: primitive}.

Consequently, based on the above designs, we can directly modify the architecture of the baseline network to implement our compositional model, by replacing $f(x) \in R^{1 \times d}$ with $X \in R^{n \times d}$ where $n = S$, and replacing $W \in R^{|Y_0| \times d}$ with $Z \in R^{|Y_0| \times N \times d}$.

\subsection{Set Similarity Function Design}

Next, we need to design the similarity function $sim(\cdot,\cdot)$, with a crucial issue to find matches between two sets. A straightforward way is to enumerate all matches as:

\vspace{-0.5cm}
{\small
	\begin{align}
		\frac{1}{nN} \sum_{i, k}^{n, N} \frac{X_i}{||X_i||} \frac{Z^y_k}{||Z^y_k||} = (\frac{1}{n} \sum_i^n \frac{X_i}{||X_i||}) (\frac{1}{N} \sum_k^N \frac{Z^y_k}{||Z^y_k||}),
		\label{eq:baseline_match}
\end{align}}where $Z^y \in R^{N \times d}$ and the cosine similarity is used as $s(\cdot,\cdot)$ to measure the similarity between primitives. 

However, Eq.\ref{eq:baseline_match} indicates that this strategy degenerates the separated primitives into the averaged feature $\frac{1}{n} \sum_i^n \frac{X_i}{||X_i||}$ which closely resembles $f(x) = \frac{1}{S} \sum_{i}^{S} F(x)_i$. This approach lacks the flexibility to capture the compositional information inherent in each class and sample. Furthermore, it may incorporate sample-specific candidate primitives, such as the background, into the primitive set, because the matching score between these sample-specific candidate primitives and $Z^y$ could be mistakenly high.

Therefore, we set a weight for each matching as

\vspace{-0.3cm}
{\small
	\begin{equation}
		sim(X, Z^y) = \frac{1}{nN} \sum_{i,k} w^A_{ik} \frac{X_i}{||X_i||} \frac{Z^y_k}{||Z^y_k||}
		\label{eq:A_w}
\end{equation}}where $W^A = \{{w^A_{ik}}\}^{N \cdot n}_{ik} \in R^{n \times N}$ is a weight matrix, which filters out sample-specific candidate primitives in $X$ by assigning low weights to their similarity with the primitive set, and highlights important ones.

\textbf{Inspiration from Representation Comparison}


To obtain $W^A$,
we draw inspiration from a related field: comparing representations across different models~\cite{kornblith2019similarity}. This area focuses on comparing outputs from various neural networks to study the behavior of deep models, typically based on features extracted from the same set of images~\cite{kornblith2019similarity}.
However, as representations extracted by different models may not share the same set of channels, simple calculations like Euclidean or cosine similarity may not be effective. 
Therefore, various approaches have been proposed to handle unmatched channels, including linear regression\cite{romero2015fitnets}, CCA\cite{raghu2017svcca}, SVCCA\cite{raghu2017svcca}, DeepEMD\cite{oh2022understanding}, etc. Among these, CKA\cite{kornblith2019similarity} shows better reliability and lower computational cost\cite{davari2022reliability}.

To compare representations of different models (RC), \textbf{the same batch of inputs} is employed, but \textbf{channels are disordered} and challenging to compare directly.
In the comparison between $X$ and $Z^y$, (candidate) primitives are described by \textbf{the same set of channels}, but \textbf{primitives are disordered} and hard to be compared directly. 
Such symmetry inspires us to \textit{view the channel dimension in RC as the primitive dimension in $sim(X, Z^y)$ and the batch dimension in RC as the channel dimension in $sim(X, Z^y)$}. Therefore, we propose to use CKA as a better similarity function to obtain $W^A$ in Eq.\ref{eq:A_w} by simply transposing $X$ and $Z^y$.
In summary, CKA captures the correlation between channels given an ordered batch, and we propose to use it to capture the correlation between primitives given ordered channels.

Specifically, in RC, given two models $h(\cdot)$ and $g(\cdot)$ for comparison, features are extracted from $X^r$ with the batch size $b^r$ as $h(X^r) \in R^{b^r \times d_h}$ and $g(X^r) \in R^{b^r \times d_g}$ to obtain the CKA similarity as 

\vspace{-0.5cm}
{\small
	\begin{align}
		&{\rm CKA} = \frac{{\rm HSIC}(h(X^r), g(X^r))}{\sqrt{{\rm HSIC}(h(X^r), h(X^r)) \cdot {\rm HSIC}(g(X^r), g(X^r))}}, \nonumber\\
		&{\rm HSIC}(K, L) = \frac{1}{(b^r-1)^2} tr(KHLH)
		\label{eq:CKA_RC}
\end{align}}where $K$ and $L$ denotes $h(X^r)$ and $g(X^r)$, $H$ is the centering matrix $H_n = I_n - \frac{1}{n} \boldsymbol{1} \boldsymbol{1}^\top $~\cite{kornblith2019similarity}.

For our $sim(X, Z^y)$, we follow Eq.\ref{eq:CKA_RC} to replace $h(X^r)$ and $g(X^r)$ with $X^\top$ and ${Z^y}^\top$ respectively. Then, the similarity between $X$ and $Z^y$ can be obtained as 

\vspace{-0.5cm}
{\small
	\begin{align}
		sim(X, Z) &= \frac{{\rm HSIC}(X^\top, {Z^y}^\top)}{\sqrt{{\rm HSIC}(X^\top, {Z^y}^\top) \cdot {\rm HSIC}(X^\top, {Z^y}^\top)}} \\
		&\overset{\text{linear}}{=}
		\frac{||\tilde{X} \tilde{Z^y}^\top||^2}{||\tilde{X} \tilde{X}^\top|| \cdot ||\tilde{Z} \tilde{Z^y}^\top||}.
		\label{eq:CKA_SC}
\end{align}}Recent studies have demonstrated that the linear CKA similarity exhibits reliability comparable to the kernel CKA similarity. Hereafter, we employ the linear CKA similarity as our CKA-based similarity function to approximate $|Z^y \cap X|$ in Eq.\ref{eq:set_similarity}.
In the linear CKA, Eq.\ref{eq:CKA_SC} is expressed equivalently with the centered features $\tilde{X} = X - \frac{1}{d} \sum_j X_{:,j}$ and $\tilde{Z^y} = Z^y - \frac{1}{d} \sum_j Z^y_{:,j}$, where the dot product is computed in the channel dimension instead of the primitive dimension.

\textbf{Benefiting Compositions with Less Computational Overhead and Robustness to Primitive Noises}

By expanding Eq.\ref{eq:CKA_SC}, we have

\vspace{-0.5cm}
{\small
	\begin{align}
		&sim(X, Z^y) = \sum_{i, k}^{n, N} \frac{ (\tilde{X}_i \tilde{Z^y}_k^\top) }{||\tilde{X} \tilde{X}^\top|| \cdot ||\tilde{Z^y} \tilde{Z^y}^\top||} \cdot (\tilde{X}_i \tilde{Z^y}_k^\top) \\
		&= {\scriptsize
		\sum_{i, k}^{n, N} \underbrace{\frac{ (\frac{\tilde{X}_i}{||\tilde{X}_i||} \frac{\tilde{Z^y}_k}{||\tilde{Z^y}_k||}^\top) }{||\frac{\tilde{X}}{||\tilde{X}||} \frac{\tilde{X}}{||\tilde{X}||}^\top|| \cdot ||\frac{\tilde{Z^y}}{||\tilde{Z^y}||} \frac{\tilde{Z^y}}{||\tilde{Z^y}||}^\top||} }_{(w^A_{ik})} \cdot \underbrace{(\frac{\tilde{X}_i}{||\tilde{X}_i||} \frac{\tilde{Z^y}_k}{||\tilde{Z^y}_i||}^\top)}_{(\frac{X_i}{||X_i||} {\frac{Z^y_k}{||Z^y_k||}}^\top)}}
		\label{eq:A_c}
\end{align}}where $||\tilde{X}||$ and $||\tilde{Z}||$ denotes the row-wise norm.

Compare Eq.\ref{eq:A_w} and Eq.\ref{eq:A_c}, we can see 
Eq.\ref{eq:A_c} well matches Eq.\ref{eq:A_w} by automatically generating the $W^A_{ik}$ and replacing $X$ and $Z$ with $\tilde{X}$ and $\tilde{Z}$ respectively.
Since $W^A_{ik}$ is obtained through the matrix multiplication, no extra computations 
are needed, \textbf{reducing the computational overhead} compared with DeepEMD or FRN\cite{wertheimer2021few}. 

Moreover, since DeepEMD or FRN obtain the matching weight by globally taking all patches into account, it makes them vulnerable to primitive noises. Such noise exists for two reasons: (1) each sample contains sample-specific candidate primitives, but the features for them are not well trained, due to their marginal contribution to classification; (2) features are not discriminative enough at the early of training. Such noises would make the complex linear programming in DeepEMD or FRN fragile, harming the set comparison.
In contrast, in Eq.\ref{eq:A_c}, each matching weight is generated by mainly taking the local comparison of two patches ($\tilde{X}_i$ and $\tilde{Z}^y_k$). Such simplicity makes the comparison less vulnerable to noisy patch features and more robust.

To verify the computational efficiency and the noise robustness, we train a model with only the baseline classification loss on CIFAR100, and then conduct an evaluation based on CKA, DeepEMD, and FRN. The average time and accuracy of the last session are reported in Tab.\ref{tab:cka-emd-recon}. Since the model is not trained with the corresponding distance metric, the feature extracted by it could be understood to be ineffective and thus noisy. We can see CKA shows the highest performance under such noisy features with the lowest time cost.

\begin{table}[t]
	\begin{center}
		\caption{Evaluation of models trained by the baseline method. CKA and the power transformed CKA shows less computational overhead and better robustness to primitive noises.}
		\label{tab:cka-emd-recon}
		\resizebox{0.48\textwidth}{!}{
			\begin{tabular}{lccc|c}
				\toprule
				& FRN & DeepEMD & CKA & Power transformed CKA
				\\
				\midrule
				Last session accuracy (\%) $\uparrow$ & 18.15 & 24.03 & 38.42 & \textbf{39.47} \\
				Time (sec. / 100 images) $\downarrow$ & 0.0233 & 12.3166 & \textbf{0.0139} & 0.0161 \\
				\bottomrule
			\end{tabular}
		}
	\end{center}
\end{table}

As CKA is robust to patch noises, we further propose to apply a power transformation on it to enhance such robustness.
Specifically, we introduce the transformation on feature maps by replacing $X_i$ with $X_i^\alpha$ element-wisely, where $\alpha < 1.0$. This action smooths the distribution of matching weights to avoid outliers caused by primitive noises. 



Based on Eq.\ref{eq:compose_to_recognize} and Eq.\ref{eq:A_c}, we can classify $x$ by trying to compose it with primitive sets from different classes, which also brings the training loss as

\vspace{-0.5cm}
{\small
	\begin{align}
		L_{cmp} = -ln \frac{e^{\frac{sim(X^\alpha, Z^y)}{|X|}}}{\sum_k e^{\frac{sim(X^\alpha, Z^k)}{|X|}}} = -ln \frac{e^{\tau \cdot sim(X^\alpha, Z^y)}}{\sum\limits_k e^{\tau \cdot sim(X^\alpha, Z^k)}}
		\label{eq:L_cmp}
\end{align}}where $\tau$ is a temperature parameter to absorb $|X|$ since $|X|$ is a constant. 
Since we have $|Z^y \cap X| \approx sim(X, Z^y)$ in Eq.\ref{eq:set_similarity}, 
we call $sim(X^\alpha, Z^y)$ the composition score.

\subsection{Primitive Reuse Design}

\begin{table*}[t]
	\caption{Comparison on the \textit{mini}ImageNet dataset. PD: lower performance drop indicates less forgetting.}
	\label{tab:mini_sota}
	\centering
	\resizebox{0.85\textwidth}{!}{
		\begin{tabular}{llcccccccccc}
			\toprule
			Backbone & Method & S0 & S1 & S2 & S3 & S4 & S5 & S6 & S7 & S8 & PD $\downarrow$\\
			\midrule
			\multirow{6}{*}{ResNet18} 
			& DeepEMD\cite{zhang2020deepemd} & 69.77 & 64.59 & 60.21 & 56.63 & 53.16 & 50.13 & 47.79 & 45.42 & 43.41 & 26.36\\
			& CLOM\cite{zou2022margin} & 73.08 & 68.09 & 64.16 & 60.41 & 57.41 & 54.29 & 51.54 & 49.37 & 48.00 & 25.08 \\
                & SoftNet\cite{SoftNet} & 76.63 & 70.13 & 65.92 & 62.52 & 59.49 & 56.56 & 53.71 & 51.72 & 50.48 & 26.15 \\
			& ALICE\cite{Peng2022few} & 80.60 & 70.60 & 67.40 & 64.50 & 62.50 & 60.00 & 57.80 & 56.80 & 55.70 & 24.90\\
			& SAVC\cite{Song_2023_CVPR} & 81.12 & 76.14 & 72.43 & 68.92 & 66.48 & 62.95 & 59.92 & 58.39 & 57.11 & 24.01\\
			\cmidrule{2-12}
			& Comp-FSCIL & \textbf{82.78} & \textbf{77.82} & \textbf{73.70} & \textbf{70.57} & \textbf{68.26} & \textbf{65.11} & \textbf{62.19} & \textbf{60.12} & \textbf{59.00} & \textbf{23.78} \\
			\midrule
			\multirow{2}{*}{ResNet12} & NC-FSCIL\cite{Yang2023Neural} & \textbf{84.02} & 76.80 & 72.00 & 67.83 & 66.35 & 64.04 & 61.46 & 59.54 & 58.31 & 25.71 \\
			\cmidrule{2-12}
			& Comp-FSCIL & 84.00 & \textbf{78.49} & \textbf{74.44} & \textbf{71.51} & \textbf{69.30} & \textbf{66.61} & \textbf{63.66} & \textbf{61.64} & \textbf{60.61} & \textbf{23.39} \\
			\bottomrule
		\end{tabular}
	}\vspace{-0.3cm}
\end{table*}

\begin{table}[t]
	\begin{center}
		\caption{Dataset~\cite{zhang2021few}. Every dataset provides fixed training and test sets, so the sampling of episodes is not needed.}
		\label{tab:dataset}
		\resizebox{0.48\textwidth}{!}{
			\begin{tabular}{lcccccc}
				\toprule
				Dataset & Total & Base  & Novel & Inc. Sessions & Shot & Image Size
				\\
				\midrule
				\textit{mini}ImageNet & 100 & 60 & 40 & 8 & 5 & 84 $\times$ 84 \\
				CUB200 & 200 & 100 & 100 & 10 & 5 & 224 $\times$ 224 \\
				CIFAR100 & 100 & 60 & 40 & 8 & 5 & 32 $\times$ 32 \\
				\bottomrule
			\end{tabular}
		}
	\end{center}\vspace{-0.4cm}
\end{table}

Primitives are shared among classes, enhancing the interpretability of the compositional model. However, in the current design, primitive sets are learned independently in each class, restricting the reuse of primitives. To address this limitation, we introduce a primitive-reuse module to enhance the correlation between primitives across classes.

During the base-session training, given the primitive set $Z^y = \{P^y_i\}_i^N$ from the class $y$, we use primitives from other base classes to replace $Z^y$.
Denote primitives from other base classes as $\{P^o_k\}_k^{N \cdot (|Y_0| - 1)}$, for each $P^y_i$, we obtain its similarity with other primitives as $s_r(P^y_i, P^o_k) = - ||P^y_i - P^o_k||^2$. Then we calculate the attention on $P^o_k$ against all other base-class primitives as 

\vspace{-0.4cm}
{\small
	\begin{equation}
		att^{y,i}_{k} = \frac{e^{\gamma \cdot s_r(P^y_i, P^o_k)}}{\sum_{k=1}^{N \cdot (|Y_0| - 1)} e^{\gamma \cdot s_r(P^y_i, P^o_k)} },
		\label{eq:atten}
\end{equation}}where $\gamma$ is a pre-defined hyper-parameter.
The replacement is then calculated as a weighted sum of all primitives as 

\vspace{-0.3cm}
{\small
	\begin{equation}
		\hat{P^y_i} = \sum\nolimits_{k=1}^{N \cdot (|Y_0| - 1)} att^{y,i}_{k} P^o_k.
		\label{eq:recon}
\end{equation}}By setting a large $\gamma$ (e.g., 64), we push the model to focus on only the closest primitive from other classes. Then, we use $\hat{P^y_i}$ to replace $P^y_i$.
The above replacement will be carried out on all primitives $Z \in R^{|Y_0| \times N \times d}$ to obtain the replaced primitive sets $\hat{Z}$.
Finally, a classification loss will be applied to the input sample based on the replaced primitive set $\hat{Z}$ as

\vspace{-0.3cm}
{\small
	\begin{equation}
		L_{rcmp} = -ln \frac{e^{\tau \cdot sim(X^\alpha, \hat{Z^y})}}{\sum_i e^{\tau \cdot sim(X^\alpha, \hat{Z^i})}},
		\label{eq:recon_cls}
\end{equation}}where $\tau$ is a temperature parameter. During training, as $X$ can be effectively classified by the original primitive sets, minimizing $L_{rcmp}$ pushes the model to generate optimal replacements for each primitive $P^y_i$ by reducing its distance with the nearest primitives from different classes, facilitating the reuse of primitives across classes.

In the incremental session, primitives from all base classes are employed to replace novel-class primitive sets.

\subsection{Model Training and Evaluation}

During the base session, we incorporate the baseline classification loss to ensure the stability of model training. The ultimate model encompasses two classifiers: one for the ordinary classification (with parameters $W \in R^{|Y| \times d}$) and another for the compositional classification (with parameter $Z \in R^{|Y| \times N \times d}$).
In the baseline classification loss, we utilize the standard feature (i.e., the global-average-pooling feature or the CLS token feature, $f(x)$) to compute the loss

\vspace{-0.3cm}
{\small
	\begin{equation}
		L_{cls} = -ln \frac{e^{\tau \cdot s(f(x), W_y)}}{\sum_i e^{\tau \cdot s(f(x), W_i)}}.
		\label{eq:ori_cls}
\end{equation}}\vspace{-0.4cm}

In all, the model is trained with all three losses as 

\vspace{-0.3cm}
{\small
	\begin{equation}
		L = L_{cls} + \lambda_1 L_{cmp} + \lambda_2 L_{rcmp}.
		\label{eq:all_cls}
\end{equation}}\vspace{-0.5cm}

\begin{table*}[t]
	\begin{center}
		\caption{Comparison on the CIFAR100 dataset.}
		\label{tab:cifar_sotas}
		\centering
		\resizebox{0.75\textwidth}{!}{
			\begin{tabular}{llcccccccccc}
				\toprule
				Backbone & Method & S0 & S1 & S2 & S3 & S4 & S5 & S6 & S7 & S8 & PD $\downarrow$\\
				\midrule
                \multirow{4}{*}{ResNet20} 
				& DeepEMD(\citeyear{zhang2020deepemd}) & 69.75 & 65.06 & 61.20 & 57.21 & 53.88 & 51.40 & 48.80 & 46.84 & 44.41 & 25.34 \\
                & WaPR(\citeyear{kim2023warping}) & 74.21 & 69.96 & 65.86 & 61.92 & 58.74 & 55.79 & 53.50 & 51.51 & 49.33 & 24.88 \\
                & MetaFSCIL(\citeyear{9878925}) & 74.50 & 70.10 & 66.84 & 62.77 & 59.48 & 56.52 & 54.36 & 52.56 & 49.97 & 24.53 \\
                \cmidrule{2-12}
                & Comp-FSCIL & \textbf{76.00} & \textbf{71.75} & \textbf{67.67} & \textbf{63.76} & \textbf{60.99} & \textbf{57.98} & \textbf{55.98} & \textbf{54.09} & \textbf{51.61} & \textbf{24.39} \\
                \midrule
                \multirow{4}{*}{ResNet18}
                & SoftNet(\cite{SoftNet}) & 72.62 & 67.31 & 63.05 & 59.39 & 56.00 & 53.23 & 51.06 & 48.83 & 46.63 & 25.99 \\
				& ALICE(\citeyear{Peng2022few}) & 79.00 & 70.50 & 67.10 & 63.40 & 61.20 & 59.20 & 58.10 & 56.30 & 54.10 & 24.90 \\
                & WaPR(\citeyear{kim2023warping}) & 80.31 & 75.86 & 71.87 & 67.58 & 64.39 & 61.34 & 59.15 & 57.10 & 54.74 & 25.57 \\
                \cmidrule{2-12}
        		& Comp-FSCIL & \textbf{80.93} & \textbf{76.52} & \textbf{72.69} & \textbf{68.52} & \textbf{65.50} & \textbf{62.62} & \textbf{60.96} & \textbf{59.27} & \textbf{56.71} & \textbf{24.22} \\
                \midrule
                \multirow{2}{*}{ResNet12}
				& NC-FSCIL(\citeyear{Yang2023Neural}) & \textbf{82.52} & 76.82 & 73.34 & 69.68 & 66.19 & 62.85 & 60.96 & 59.02 & 56.11 & 26.41 \\
                \cmidrule{2-12}
				& Comp-FSCIL & 82.30 & \textbf{78.58} & \textbf{74.47} & \textbf{70.27} & \textbf{67.29} & \textbf{64.49} & \textbf{62.78} & \textbf{61.38} & \textbf{59.05} & \textbf{23.25} \\
				\bottomrule
			\end{tabular}
		}\vspace{-0.3cm}
	\end{center}
\end{table*}

In the incremental session, we fix the backbone network and only train the novel-class primitive set $Z^{novel}$ by Eq.\ref{eq:all_cls}. The base-class primitives are reused in novel classes, both implicitly by the transferring of the backbone network and explicitly by $L_{rcmp}$. 
During this period, the model learns the composition of reused primitives by training $Z^{novel}$ suitable for composition.
Finally, the model will be deployed to classify all encountered classes based on all primitive sets using the composition score.

\section{Experiments}


\subsection{Dataset and Implementation Details}

\begin{table*}[t]
	\caption{Comparison with state-of-the-art works on the CUB200 dataset. PD: lower performance drop indicates less forgetting.}
	\label{tab:cub_sota}
	\centering
	\resizebox{1.0\textwidth}{!}{
		\begin{tabular}{clcccccccccccc}
			\toprule
			Backbone & Method & S0 & S1 & S2 & S3 & S4 & S5 & S6 & S7 & S8 & S9 & S10 & PD $\downarrow$ \\
			\midrule
			\multirow{6}{*}{ResNet18} 
			& D-DeepEMD~\cite{zhang2020deepemd} & 75.35 & 70.69 & 66.68 & 62.34 & 59.76 & 56.54 & 54.61 & 52.52 & 50.73 & 49.20 & 47.60 & 27.75 \\
			& MetaFSCIL~\cite{chi2022metafscil} & 75.90 & 72.41 & 68.78 & 64.78 & 62.96 & 59.99 & 58.30 & 56.85 & 54.78 & 53.82 & 52.64 & 23.26 \\
                & SoftNet\cite{SoftNet} & 78.07 & 74.58 & 71.37 & 67.54 & 65.37 & 62.60 & 61.07 & 59.37 & 57.53 & 57.21 & 56.75 & 21.32 \\
                & WaPR\cite{kim2023warping} & 77.74 & 74.15 & 70.82 & 66.90 & 65.01 & 62.64 & 61.40 & 59.86 & 57.95 & 57.77 & 57.01 & 20.73 \\
                & GKEAL\cite{Zhuang_2023_CVPR} & 78.88 & 75.62 & 72.32 & 68.62 & 67.23 & 64.26 & 62.98 & 61.89 & 60.20 & 59.21 & 58.67 & 20.21 \\
			& NC-FSCIL~\cite{Yang2023Neural} & 80.45 & 75.98 & 72.30 & 70.28 & 68.17 & 65.16 & 64.43 & 63.25 & 60.66 & 60.01 & 59.44 & 21.01 \\
			& CLOM~\cite{zou2022margin} & 79.57 & 76.07 & 72.94 & 69.82 & 67.80 & 65.56 & 63.94 & 62.59 & 60.62 & 60.34 & 59.58 & 19.99 \\
			\cmidrule{2-14}
			& Comp-FSCIL & \textbf{80.94} & \textbf{77.51} & \textbf{74.34} & \textbf{71.00} & \textbf{68.77} & \textbf{66.41} & \textbf{64.85} & \textbf{63.92} & \textbf{62.12} & \textbf{62.10} & \textbf{61.17} & \textbf{19.77} \\
			\midrule
			\multirow{2}{*}{Swin-T} & CLOM~\cite{zou2022margin} & 86.28 & 82.85 & 80.61 & 77.79 & 76.34 & 74.64 & 73.62 & 72.82 & 71.24 & 71.33 & 70.50 & 15.78 \\
			\cmidrule{2-14}
			& Comp-FSCIL & \textbf{87.67} & \textbf{84.73} & \textbf{83.03} & \textbf{80.04} & \textbf{77.73} & \textbf{75.52} & \textbf{74.32} & \textbf{74.55} & \textbf{73.35} & \textbf{73.15} & \textbf{72.80} & \textbf{14.87} \\
			\bottomrule
		\end{tabular}
	}\vspace{-0.3cm}
\end{table*}

\begin{table*}
	\begin{center}
		\caption{Ablation study of modules on the last incremental session of three datasets. }
		\label{tab:ablation}
		\resizebox{0.8\textwidth}{!}{
			\begin{tabular}{lccc|ccc|ccc|ccc}
				\toprule
				\multirow{2}{*}{\tabincell{c}{Method}} & \multicolumn{3}{c}{CUB200} & \multicolumn{3}{c}{CUB200 (Swin-T)} & \multicolumn{3}{c}{CIFAR100} & \multicolumn{3}{c}{\textit{mini}ImageNet} \\
				\cmidrule{2-13}
				& Overall & Base & Novel & Overall & Base & Novel & Overall & Base & Novel & Overall & Base & Novel \\
				\midrule
				Baseline    & 57.18 & 79.48 & 45.67 & 69.18 & 85.65 & 61.71 & 53.98 & 79.92 & 44.07 & 57.72 & 82.53 & 42.82 \\
				+ Composition	& 59.25 & 80.13 & 49.01 & 71.42 & 87.08 & 63.75 & 55.30 & 81.43 & 47.52 & 58.84 & 82.85 & 44.37 \\
				+ Reusing & \textbf{61.17} & \textbf{81.06} & \textbf{51.40} & \textbf{72.80} & \textbf{87.79} & \textbf{65.12} & \textbf{59.05} & \textbf{82.30} & \textbf{51.05} & \textbf{60.61} & \textbf{84.00 }& \textbf{46.52} \\
				\bottomrule
			\end{tabular}
		}
	\end{center}
\end{table*}


Datasets are listed in Tab.\ref{tab:dataset}.
Our method is based on the code of CEC~\cite{zhang2021few}. For \textit{mini}ImageNet, we follow NC-FSCIL~\cite{Yang2023Neural} to utilize ResNet12~\cite{he2016deep} as the backbone network, and we set $\lambda_1 = \lambda_2 = 2.0$, $\alpha = 0.8$. For CIFAR100, we follow NC-FSCIL~\cite{Yang2023Neural} to utilize ResNet12 as the backbone network, and we remove the pooling operation for the first two residual blocks following ResNet20 used in \cite{zhang2021few}, to keep the spatial resolution of the output map. We set $\lambda_1 = \lambda_2 = 2.0$, $\alpha = 0.6$. For CUB200, we follow CLOM~\cite{zou2022margin} to scale the learning rate of the backbone network to 10\% of that in the FC layer, due to the pretraining from ImageNet following~\cite{zhang2021few}. We set $\lambda_1 = \lambda_2 = 0.01$, $\alpha = 0.5$. 

\subsection{Comparison with State-of-the-Art Methods}

The comparison with state-of-the-art works is in Tab.\ref{tab:mini_sota}, \ref{tab:cifar_sotas} and \ref{tab:cub_sota}, with all sessions in the incremental learning. 
From these tables, we can see that we consistently outperform current works by over 1.5\% in terms of the last session's performance, where all classes are taken into account. 
Moreover, we utilize the Swin Transformer~\cite{liu2021Swin} (the tiny version, denoted as Swin-T) as an example to evaluate our method given the pretraining of Large Vision Model (LVM, ImageNet1k in our experiments). For a fair comparison, experiments of Transformers are conducted on CUB200 where the ImageNet pretraining is utilized by other works. To compare with current works, we implement CLOM~\cite{zou2022margin} as the state-of-the-art method that has the highest last-session accuracy in Tab.\ref{tab:cub_sota}. We can still outperform it by 2.0\% in terms of the last-session accuracy. PD denotes the Performance Drop. It means the first session's accuracy subtracts the last session's accuracy, with lower values indicating less forgetting. We can also achieve the least forgetting due to the reuse of primitives.

\begin{figure}[t]
	\centering
	\includegraphics[width=0.9\linewidth]{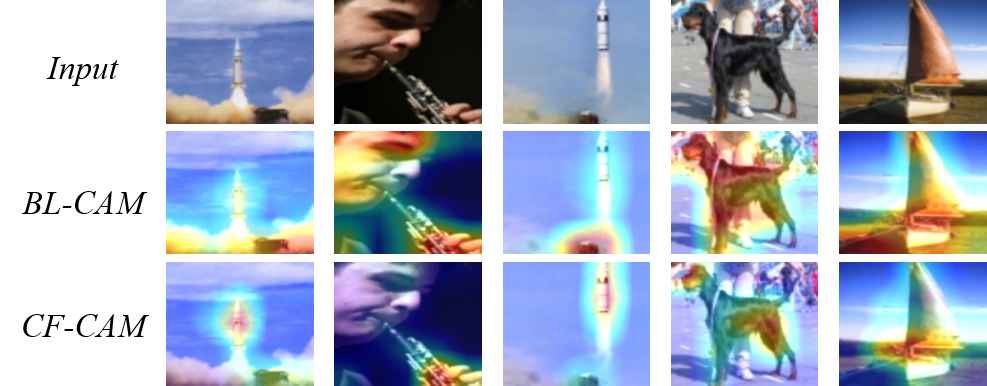}\vspace{-0.1cm}
	\caption{Visualization of class-activation-map (CAM). BL: Baseline model; CF: Our compositional model. CF-CAM activates smaller regions than BL-CAM and filters out sample-specific regions such as background, validating the focus on shared patches.}
	\label{fig:visualization}
\end{figure}

\subsection{Ablation Study}

\begin{figure}[t]
	\centering
	\hspace*{-0.3cm}
	\includegraphics[width=1.05\linewidth]{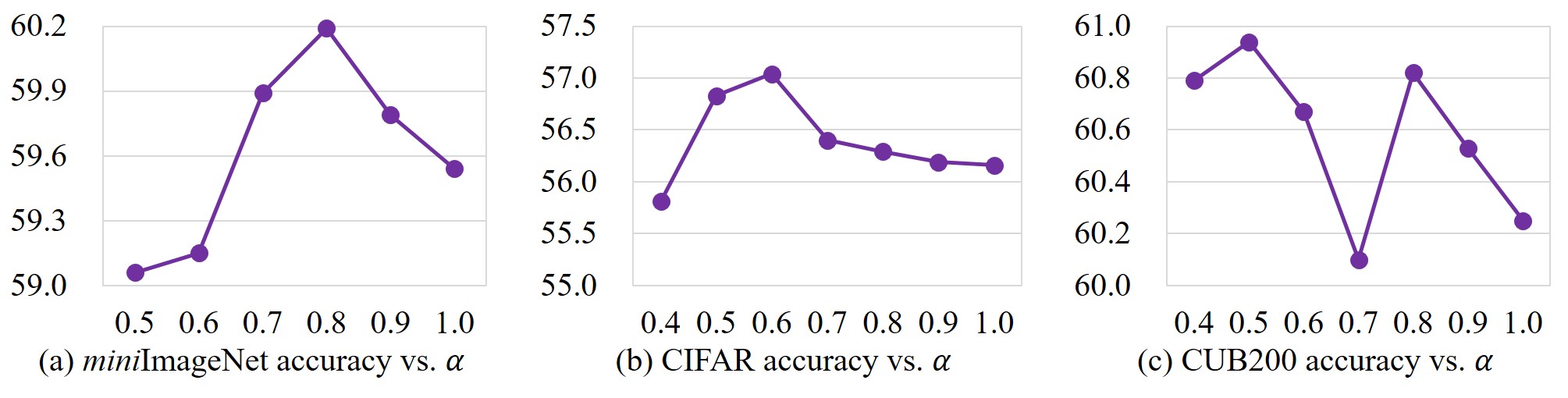}\vspace{-0.8cm}
	\caption{Sensitivity study of the power transformation parameter $\alpha$, which further improves model robustness to primitive noises.}
	\label{fig:alpha}
\end{figure}

The ablation study is reported in Tab.\ref{tab:ablation}. We include the performance of the \textit{Overall} accuracy, referring to the last-session accuracy; the \textit{Base} accuracy, referring to the base-session accuracy (S0); and the \textit{Novel} accuracy, referring to the accuracy of classifying all novel-class samples into novel classes, equivalent to the $k$-way $n$-shot evaluation in few-shot learning.
We can see each module has its contribution for all training scenarios and all performance measurements, by means of avoiding forgetting via maintaining learned primitives and fast learning of compositions.
We also report the sensitivity study of the power transformation parameter $\alpha$ in Fig.\ref{fig:alpha}, indicating it could further enhance model robustness to primitive noises.




\subsection{Primitive Effectiveness}

\subsubsection{Quantitative Analysis}

\begin{figure}[t]
	\centering
	\includegraphics[width=0.5\textwidth]{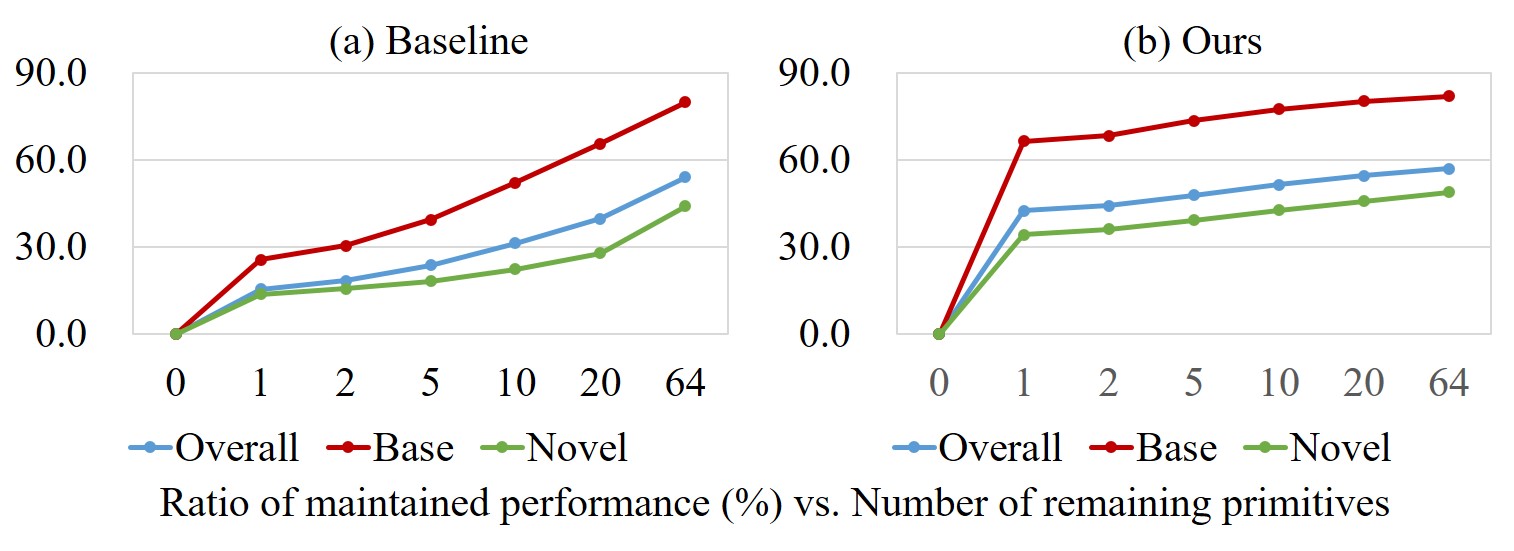}\vspace{-0.5cm}
	\caption{Quantitatively validating our method focuses more on important primitives by only maintaining them on CIFAR100.}\vspace{-0.2cm}
	\label{fig:important_primitives}
\end{figure}

Quantitatively, we test the recognition by throwing away sample-specific candidate primitives through $W^A$, and report the performance against the number of remaining primitives in Fig.\ref{fig:important_primitives}. We can see our model achieves higher performance given the same number of remaining primitives, indicating our model focuses more on important patches.

\begin{figure}[t]
	\centering
	\includegraphics[width=0.9\linewidth]{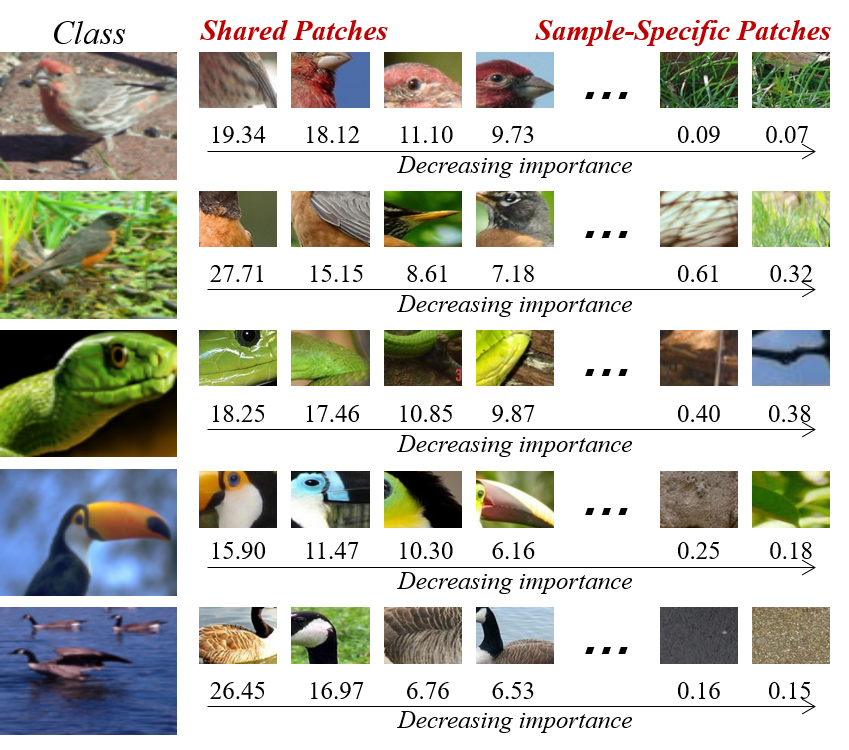}\vspace{-0.2cm}
	\caption{Image patches retrieved in each class, where important patches (candidate primitives) can represent shared patterns.}
	\label{fig:primitive}
\end{figure}

\vspace{-0.3cm}
\subsubsection{Qualitative Analysis}
\vspace{-0.1cm}

To qualitatively validate the discovered primitives, we compare the activation map of the baseline model (BL) and our compositional FSCIL model (CF), by the class-activation-map (CAM)~\cite{zhou2016learning}. Since CAM relies on the dot-product between $W_y$ and $X$ which cannot be directly applied to our CKA-based model, we rewrite the numerator in CKA as 
$\sum_i^n [\sum_k^N (\sum_j^d \tilde{X}_{ij} \tilde{Z}_{kj})^2]$ where the $[\cdot]$ denotes the designed CF-CAM.
Based on the CF-CAM visualization in Fig.\ref{fig:visualization}, we can see CF-CAM shows smaller activated regions compared with BL-CAM, which filers out sample-specific regions such as background areas. This phenomenon validates that our compositional FSCIL method could filter out sample-specific patches and highlight important (shared) ones, which improves primitives and compositions.

Then, we visualize primitives by retrieving image patches according to $W^A$, which is the importance value in Eq.\ref{eq:A_c}, for each class with samples from this class. We report the retrieved patches in Fig.\ref{fig:primitive} sorted by the importance values. We can see that candidate primitives with large importance values can indeed represent shared patterns of each class, such as furs, eyes, and beaks. In contrast, patches at the end of the sort refer to sample-specific patterns such as grass, sand, or other patterns irrelevant to the class, which are filtered out by our model through the small importance value. This phenomenon further validates that our compositional model can effectively focus on important candidate primitives and filter out sample-specific ones.


\begin{table}
	\begin{center}
		\caption{Verification of primitive reuse on CIFAR100. }\vspace{-0.2cm}
		\label{tab:reuse}
		\resizebox{1.0\linewidth}{!}{
			\begin{tabular}{cccccc}
				\toprule
				Ratio (\%) & Baseline & All Match & Max Match & + Comp & + Reuse \\
				\midrule
				1 & 100.0 & 100.0 & 100.0 & 100.0 & 100.4 \\
				2 & 99.80 & 99.69 & 99.44 & 100.0 & 100.04 \\
				5 & 99.63 & 99.56 & 99.50 & 100.04 & 100.10 \\
				10 & 98.14 & 98.14 & 98.44 & 100.14 & 100.10 \\
				20 & 93.56 & 95.72 & 96.01 & 100.35 & 101.30 \\
				50 & 82.18 & 90.83 & 92.17 & 99.02 & 99.74 \\
				80 & 69.80 & 80.42 & 82.39 & 87.86 & 89.48 \\
				90 & 62.38 & 67.28 & 70.01 & 73.14 & 76.83 \\
				\bottomrule
			\end{tabular}
		}\vspace{-0.3cm}
	\end{center}
\end{table}

\begin{figure}[t]
	\centering
	\hspace*{-0.3cm}
	\includegraphics[width=1.05\linewidth]{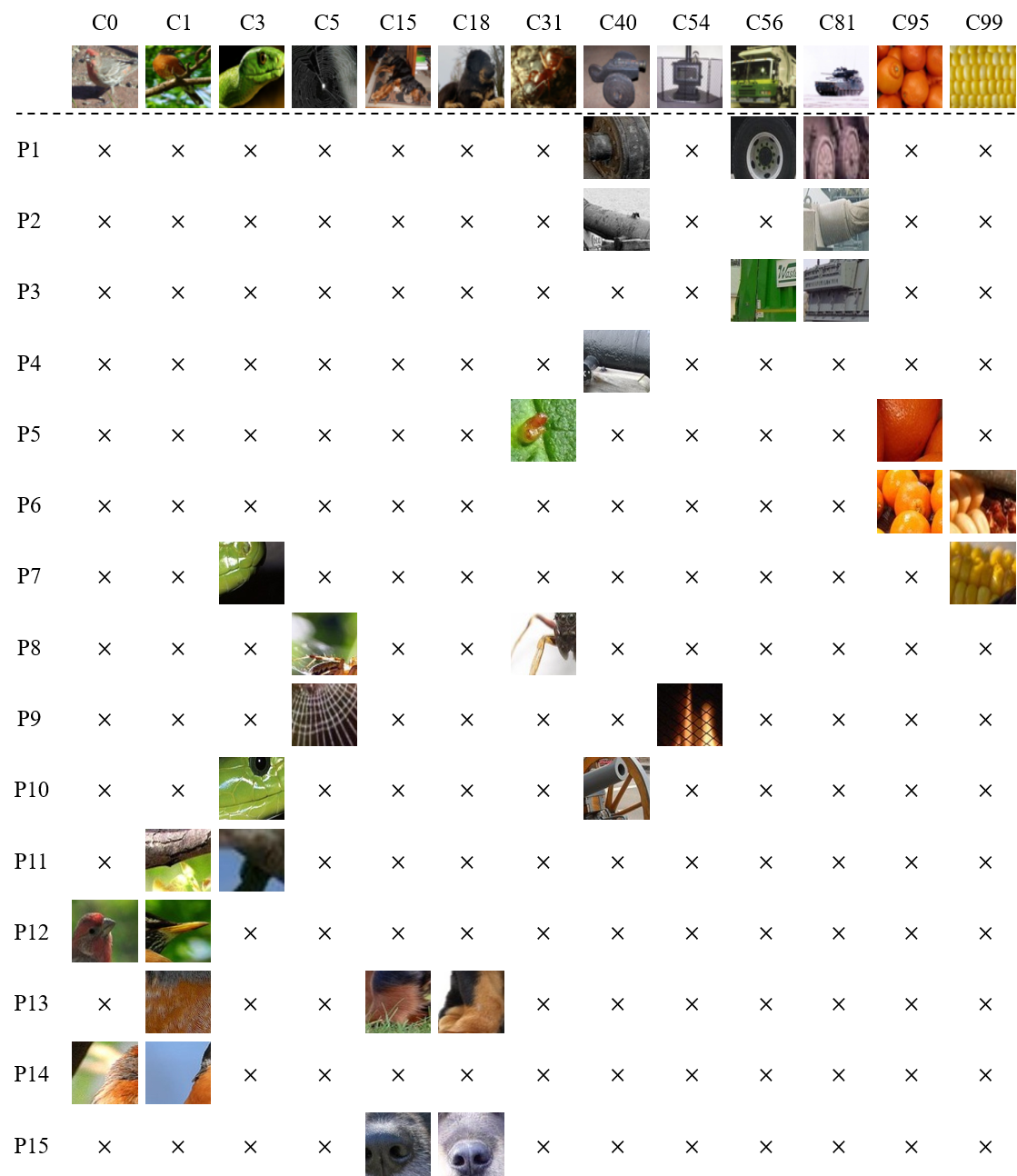}\vspace{-0.4cm}
	\caption{Image patches are retrieved across classes to validate the composition of primitives, where each row refers to a primitive. Based on it, we can interpret novel-class recognition such as \textit{The spider web is composed of a net and a spider like an ant} (P8, P9). An extended version is included in the appendix.}\vspace{-0.4cm}
	\label{fig:composition}
\end{figure}

\subsection{Primitive Reusability and Composition}

\subsubsection{Quantitative Analysis}

To verify our model can better transfer and reuse primitives than ordinary methods, we replace novel-class primitives with the nearest primitives in base classes. For the ordinary method, we use the class prototype as their primitives. We plot the ratio of replaced primitives and the ratio of remaining novel-class performance in Tab.\ref{tab:reuse}. We can see our model can achieve higher performance when primitives are replaced, validating the reusability of primitives, which lays the ground for our interpretation by primitives. 

We also compare our method with other set-similarity-based methods \cite{afrasiyabi2022MatchingFeature} in Tab.\ref{tab:reuse}. 
We can see the compositional structure (i.e., reuse of primitives) is lower than ours (the last row, when reusing ratio=90\%, the recovered ratio is much lower than ours).
Therefore, the naturally arisen reusability is far from being perfect, which needs to be strengthened by our methods.

Moreover, since in Tab.\ref{tab:reuse} the performance begin to drop only when the replace ratio reaches 80\%, we only need to re-learn 20 of primitives and can reuse other base-class primitives. Since the base session is fixed for each novel class, this means the novel-class primitive space has been compressed to 20 of its original size. These verify the potential to compress the primitive size.

\subsubsection{Qualitative Analysis}

Finally, in Fig.\ref{fig:composition}, we retrieve primitives across classes to validate the composition of primitives. We first retrieve important primitives within each class according to the importance value in each column. Then, primitives retrieved across classes with the smallest distances are in the same row. $\times$ denotes the primitive is not activated in the given class. We can see although classes are not the same, the image patches of primitives are similar, validating the reusability of primitives.
Moreover, primitives reused across classes can be viewed to compose each novel class, therefore we can interpret the recognition of each novel class in the following way:
P8 + P9: \textit{The spider web is composed of a net and a spider like an ant}.
P1 + P2 + P3: \textit{A tank is composed of a vehicle with armor and a gun barrel.}

\subsubsection{Number of Primitives}

\begin{table}
	\begin{center}
		\caption{Ablate the primitive number on CIFAR100. }
		\label{tab:primitive_number}
		\resizebox{0.78\linewidth}{!}{
			\begin{tabular}{cccc}
				\toprule
				Ratio (\%) & Base Classes & Novel Classes & All Classes \\
				\midrule
				1 & 57.20 & 34.90 & 33.20 \\
				4 & 75.01 & 40.22 & 48.32 \\
				9 & 75.35 & 41.47 & 51.55 \\
				16 & 76.00 & 42.57 & 51.61 \\
				25 & 75.25 & 42.32 & 51.49 \\
				36 & 75.50 & 42.00 & 51.28 \\
				49 & 76.16 & 41.37 & 51.23 \\
				64 & 76.18 & 42.27 & 51.60 \\
				81 & 76.28 & 41.27 & 51.06 \\
				100 & 75.76 & 41.10 & 51.37 \\
				\bottomrule
			\end{tabular}
		}\vspace{-0.4cm}
	\end{center}
\end{table}

We report experiments on CIFAR in Tab.~\ref{tab:primitive_number} to ablate primitive numbers.
Since the feature map of CIFAR is at the size of 8$\times$8, we increase the number of primitives squarely. We can see the performance reaches the top after the primitive num reaches 9 or 16, which is not a heavy burden compared with the parameters in the deep networks (e.g., 512 * 9 = 4k parameters in the primitive size vs. millions of parameters in the ResNet backbone). If the primitive size is too small, the model will lack the flexibility to represent base knowledges. If the primitive size keeps increasing, although the capacity is larger, it also imports less effective primitives and thus fails to keep improving the performance.
In our experiments, we choose 16 as the primitive size.

\section{Conclusion}

To imitate human's ability of compositional learning, we propose a compositional FSCIL method to divide knowledge into primitives and learn novel knowledge by the composition of primitives. Experiments on three datasets validate the rationale and effectiveness of our method.

\section*{Acknowledgements}
This work is supported by National Natural Science Foundation of China under grants 62206102, 62376103, 62302184, U1936108 and Science and Technology Support Program of Hubei Province under grant 2022BAA046.

\section*{Impact Statements}

We propose a cognitive-inspired method to handle the FSCIL problem by simulating humans' ability to compositional learning. This work can also be adopted in other fields like few-shot learning, and image retrieval, since the compositional structure of knowledge exists in many other domains. The limitation of this work is the neglect of the many-shot scenarios where the update of primitives cannot be ignored. However, as our method can provide a good initialization for the future update of primitives, it will also benefit the many-shot scenarios.

\bibliography{yixiongz}
\bibliographystyle{icml2024}


\appendix

\twocolumn[
\icmltitle{Appendix for Compositional Few-Shot Class-Incremental Learning}




]




\setcounter{figure}{0}   
\setcounter{table}{0}   

\section{Detailed Dataset Description}

\begin{figure}[h]
	\centering
	\includegraphics[width=0.5\linewidth]{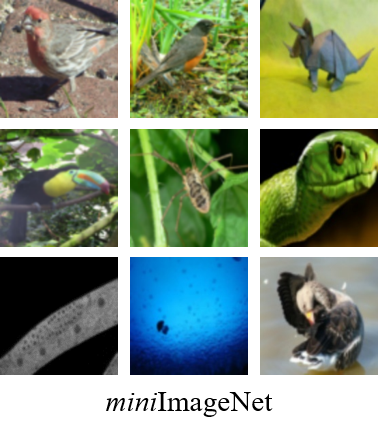}\vspace{-0.6cm}
	\caption{Samples of \textit{mini}ImageNet.}
	\label{fig:samples_mini}
\end{figure}

\textbf{\textit{mini}ImageNet}~\cite{Vinyals2016Matching} contains 100 classes with 600 samples in each class randomly sampled from ImageNet~\cite{deng2009imagenet}, which is relevant to the recognition of general objects such as cats, dogs, instruments and so on. Some samples of \textit{mini}ImageNet are shown in Fig.~\ref{fig:samples_mini}. Following current works~\cite{zhang2021few,zhou2022forward}, images are resized to 84 $\times$ 84, and 60 classes are utilized as base classes, while the remaining 40 classes are divided into 8 sessions for incremental learning, where only 5 training samples are available for each novel class.

\begin{figure}[t]
	\centering
	\includegraphics[width=0.5\linewidth]{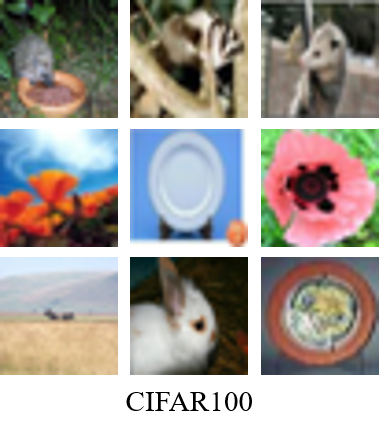}\vspace{-0.6cm}
	\caption{Samples of CIFAR100.}
	\label{fig:samples_cifar}
\end{figure}

\textbf{CIFAR100}~\cite{krizhevsky2009learning} also contains 100 classes relevant to the recognition of general objects. Samples are shown in Fig.~\ref{fig:samples_cifar}, where each image is at the size of 32 $\times$ 32. Similar to \textit{mini}ImageNet, following current works~\cite{zhang2021few,zhou2022forward}, 60 classes are selected as base classes, and the remaining 40 classes are divided into 8 incremental sessions with 5 training samples in each novel class.

\begin{figure}[h]
	\centering
	\includegraphics[width=0.5\linewidth]{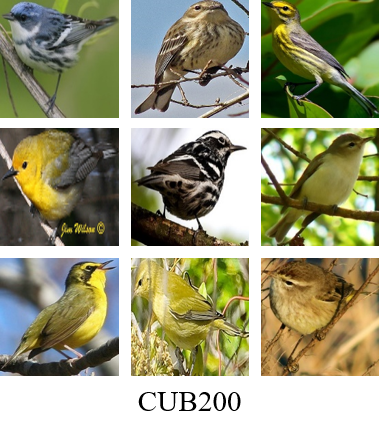}\vspace{-0.6cm}
	\caption{Samples of CUB200.}
	\label{fig:samples_cub}
\end{figure}

\textbf{CUB-200-2011 (CUB200)}~\cite{wah2011caltech} is a fine-grained dataset of birds with 200 classes in all. Samples are shown in Fig.~\ref{fig:samples_cub}, where the input size for each image is 224 $\times$ 224. Following current works~\cite{zhang2021few,zhou2022forward}, 100 classes are selected as base classes, and the remaining 100 classes are separated into 10 sessions for incremental learning.

\section{More Experiments}

\subsection{Extended Primitive Visualization}

\begin{figure*}[t]
	\centering
	\includegraphics[width=1.0\linewidth]{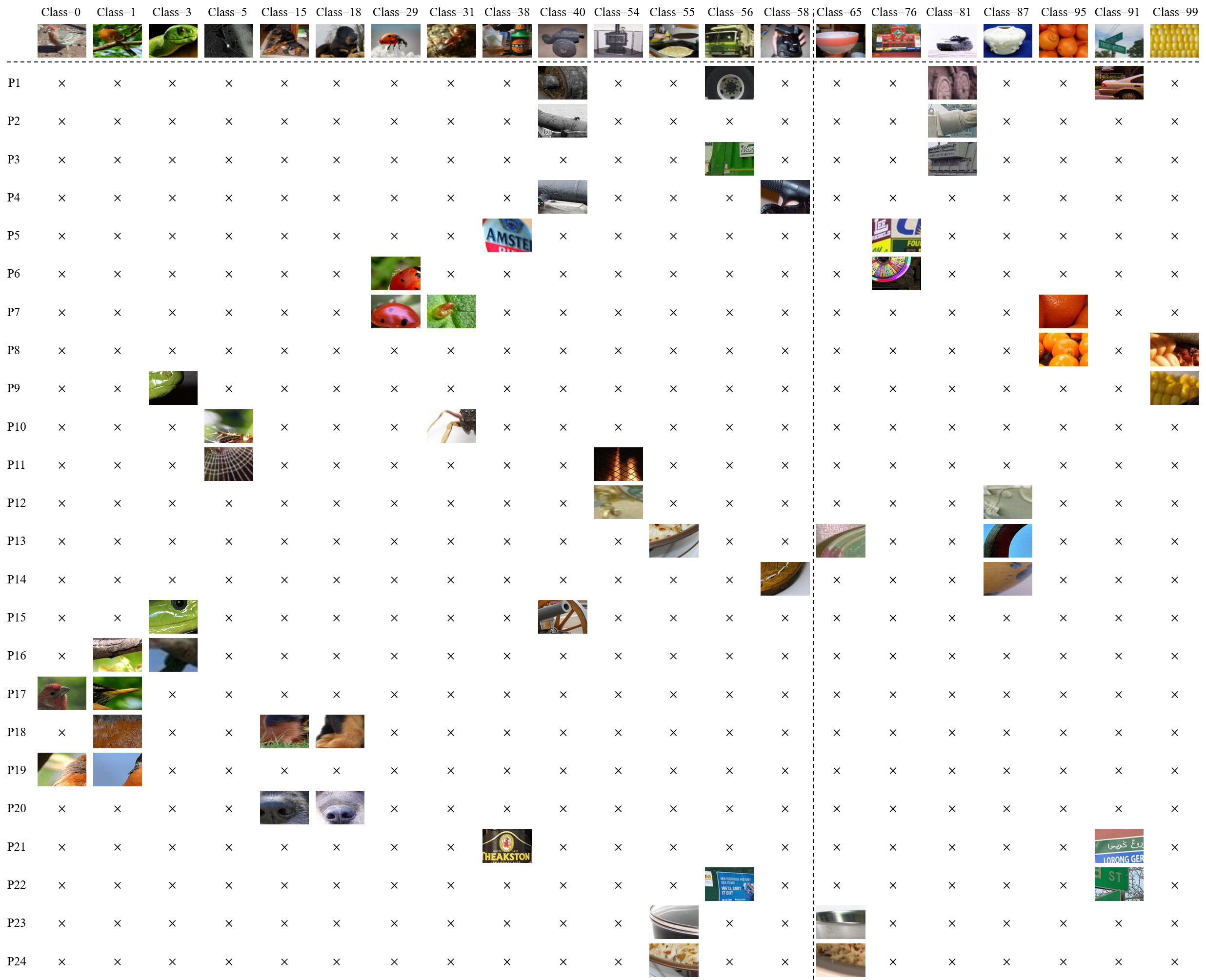}\vspace{-0.6cm}
	\caption{Extended primitive visualization across classes. Image patches are retrieved across classes to validate the composition of primitives, where each row refers to a primitive. Based on it, we can interpret novel-class recognition such as \textit{The spider web is composed of a net and a spider like an ant} (P10, P11).}
	\label{fig:composition_extend}
\end{figure*}

We provided an extended visualization of primitive across. Similar to section 4.5.2, in Fig.\ref{fig:composition_extend}, each column refers to a \textit{mini}ImageNet class, and each row refers to a primitive. We can see that primitives are reused across classes by sharing similar semantic meanings.

\begin{figure}[h]
	\centering
	\includegraphics[width=1.0\linewidth]{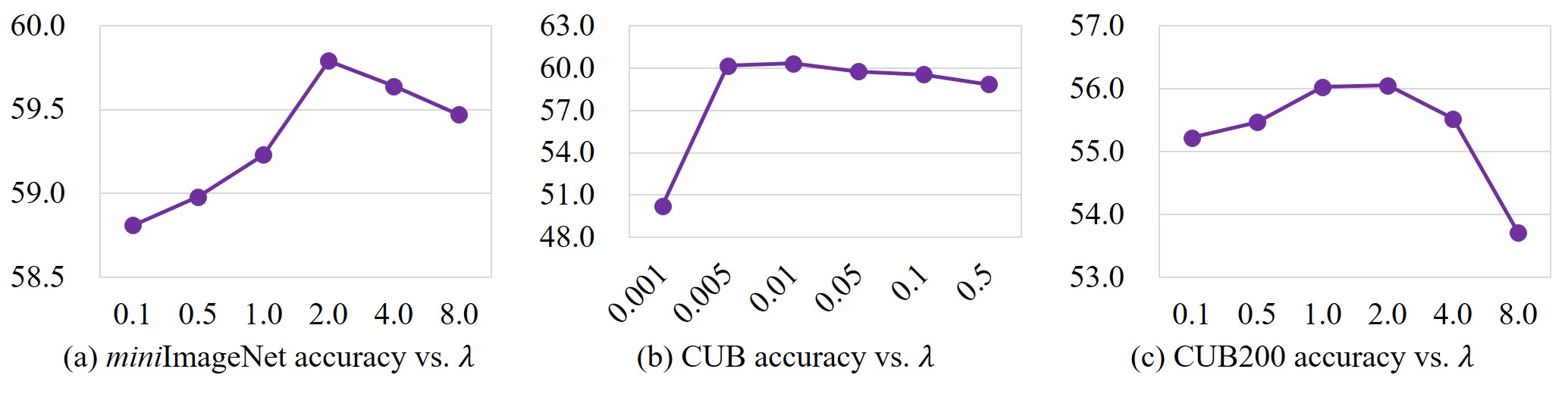}\vspace{-0.6cm}
	\caption{Sensitivity study of hyper-parameters $\lambda$.}
	\label{fig:sensitivity_lambda}
\end{figure}

\subsection{Sensitivity Study}

We also provide the sensitivity study of hyper-parameters from the primitive diversification module on CIFAR100, \textit{mini}ImageNet and CUB200 in Fig.\ref{fig:sensitivity_lambda}. We can see these three datasets show similar trends. Take CIFAR100 for an example.
$\lambda$, as the importance of the CKA similarity, achieves the highest accuracy around 2.0, meaning both the classical and CKA similarity are important in learning effective primitives. 
Moreover, on the CUB200 dataset, the optimal $\lambda$ is significantly smaller than that on the other datasets. This is because the ImageNet pretraining is utilized on CUB200, which requires the model to make better use of the pretraining. Since the pretraining is based on the classical similarity function, the weight of the loss generated by the classical similarity function should be larger. 

\subsection{Comparison with Compositional Learning Works}

To compare our work with other compositional-learning-based works, we also implemented CPDE~\cite{zou2020compositional} and RPC~\cite{mishra2022learning} on the CIFAR100 dataset following the setting provided in section 4.1. The results are reported in Tab.~\ref{tab:cifar_comp}, where we can outperform it in terms of the accuracy on each session. This is because CPDE build primitives from the aspect of channels, but we build primitives based on image patches. Although each channel can indeed represent semantic patterns, it still takes the whole image as input, which makes it vulnerable to noisy patches such as background. Moreover, the comparison between primitive sets of CPDE, however, is still modeled as the cosine similarity between every two holistic features, which can hardly prevent it from being affected by noisy patterns. In contrast, our method can efficiently filter out noisy patches and highlight important ones, which therefore benefits our model with higher performance. 
For the RPC method, this method also learns primitives from the spatial dimension. However, it forces the model to learn and recognize through a fixed dictionary of primitives, which lacks the flexibility to capture the sample-specific primitives. Therefore, our method can also outperform RPC.

\begin{table*}[t]
	\caption{Comparison with compositional learning works on the CIFAR100 dataset.}
	\label{tab:cifar_comp}
	\centering
	\resizebox{0.88\textwidth}{!}{
		\begin{tabular}{lcccccccccc}
			\toprule
			Method & S0 & S1 & S2 & S3 & S4 & S5 & S6 & S7 & S8 & PD $\downarrow$ \\
			\midrule
			CPDE~\cite{zou2020compositional}  & 80.85 & 76.09 & 71.67 & 67.69 & 64.31 & 61.49 & 59.08 & 56.79 & 54.54 & 26.31 \\
			RPC~\cite{mishra2022learning} & 80.65 & 76.22 & 72.11 & 68.04 & 64.61 & 61.93 & 59.60 & 57.41 & 55.28 & 25.37 \\
			Ours & \textbf{82.30} & \textbf{78.58} & \textbf{74.47} & \textbf{70.27} & \textbf{67.29} & \textbf{64.49} & \textbf{62.78} & \textbf{61.38} & \textbf{59.05} & \textbf{23.25} \\
			\bottomrule
		\end{tabular}
	}
\end{table*}

\section{Extended Related Work}

\textbf{Few-shot class-incremental learning} (FSCIL) can be roughly grouped into adaptation-based~\cite{hou2019learning,rebuffi2017icarl,castro2018end,tao2020few} and metric-based methods~\cite{zhang2021few,zou2022margin}. The first group adapts the model during novel-class training, but the backbone network may be frozen to avoid catastrophic forgetting~\cite{zhou2022forward}. 
For example, CEC~\cite{zhang2021few} meta-trains the the graph network for propagating the classifier information according to contexts on base classes, and then transfers the propagation mechanism to novel classes for generating novel-class classifiers. FACT~\cite{zhou2022forward} reserves feature space for novel classes to avoid the conflicts between novel classes and base classes, so as to alleviate the catastrophic forgetting brought by the novel-class finetuning.
The second group represents each class through prototypes averaged from samples~\cite{zou2022margin}, which also freezes network parameters to avoid catastrophic forgetting. 
For example, CLOM~\cite{zou2022margin} learns a margin-based feature extractor to improve the representations, and recognizes novel classes by the distance between prototypes and each sample's representation.
However, most of current works learn a holistic feature for each input sample, and seldom works studied the compositional structure of the FSCIL models.
To the best of our knowledge, we are the first to discover the compositional components of the learned knowledge, and build a compositional model with both higher performance and better interpretability.

\textbf{Compositional learning} aims to learn through primitives (components) of knowledge, which has been well studied in cognitive science~\cite{biederman1987recognition, hoffman1984parts, fodor1975language}. Some works applied this concept in other domains.
For example, CompCos~\cite{zou2020compositional} decomposes classes into channels for few-shot learning, which views the cosine similarity between prototypes and input samples as the element-wise comparison between primitive sets.
CORL~\cite{DBLP:journals/corr/abs-2101-11878} decomposes knowledge into pre-defined visual prototypes learned on base classes, and utilizes pre-defined activation maps for novel-class composition.
\cite{purushwalkam2019task} decomposes visual features to attributes for zero-shot learning, which encourages the visual features to be close to the combination of attribute features.
\cite{kato2018compositional} decomposes human-object interactions into actions and objects. 
However, seldom effort has been made for the FSCIL task so far, and most of the current works~\cite{purushwalkam2019task,kato2018compositional} rely on the extra attribute or part annotations. Compared with them, our decomposition is from the spatial dimension and does not require additional annotations for primitives.



\end{document}